%% file: main.tex
\documentclass[sigconf,screen,natbib=true]{acmart}
\copyrightyear{2026}
\acmYear{2026}
\setcopyright{cc}
\setcctype{by-nc-nd}
\acmConference[WWW '26]{Proceedings of the ACM Web Conference 2026}{April 13--17, 2026}{Dubai, United Arab Emirates}
\acmBooktitle{Proceedings of the ACM Web Conference 2026 (WWW '26), April 13--17, 2026, Dubai, United Arab Emirates}
\acmPrice{}
\acmDOI{10.1145/3774904.3792993}
\acmISBN{979-8-4007-2307-0/2026/04}

\usepackage[dvipsnames]{xcolor}
\usepackage{enumitem}
\usepackage{latexsym}
\usepackage{hyperref}
\usepackage[T1]{fontenc}
\usepackage[utf8]{inputenc}
\usepackage{bbding}
\usepackage{microtype}
\usepackage{amsmath}

\usepackage{booktabs}
\usepackage{multirow}
\usepackage{colortbl}
\usepackage{xcolor} 
\usepackage{makecell}
\usepackage{graphicx}
\usepackage{arydshln}
\usepackage{float}
\usepackage{caption}
\usepackage{subcaption}
\usepackage{verbatim}
\usepackage{tcolorbox}
\tcbuselibrary{breakable}
\usepackage{array}
\usepackage{pgfplots}
\pgfplotsset{compat=1.18}
\usepgfplotslibrary{groupplots}
\pgfplotsset{
    debate round style/.style={
        ybar,
        width=\linewidth,
        height=4cm,
        ylabel={Number of Claims},
        xlabel={Number of Debated Rounds},
        symbolic x coords={1, 2, 3},
        xtick=data,
        ymin=0,
        ymax=305,
        bar width=8pt,
        enlarge x limits=0.25,
        grid=major,
        grid style={dashed, gray!30},
        every axis plot/.append style={
            fill opacity=0.85,
            draw opacity=1,
            line width=1pt
        },
        ylabel style={font=\normalsize},
        tick label style={font=\normalsize},
        nodes near coords,
        nodes near coords align={vertical},
        every node near coord/.append style={font=\small, color=black, yshift=0.5pt}
    }
} 
\usepackage[normalem]{ulem}

\newcommand{\flaw}[1]{\textcolor{purple}{#1}}
\newcommand{\good}[1]{\textcolor{Green}{#1}}
\newcommand{\adver}[1]{\textcolor{Cyan}{#1}}

\newtcolorbox{calloutblock}{
    colback=white,
    colframe=black,
    boxrule=0.5pt,
    arc=0pt,
    left=1pt,
    right=1pt,
    top=1pt,
    bottom=1pt,
    breakable,
    fontupper=\footnotesize,
}

\begin{document}

\title{\textit{Debating Truth}: Debate-driven Claim Verification with Multiple Large Language Model Agents}

\author{Haorui He}
\affiliation{%
  \institution{Department of Interactive Media, Hong Kong Baptist University}
  \country{}
  }
\affiliation{%
  \institution{School of Computing and Data Science, The University of Hong Kong}
  \country{}
  }
\email{harryhe@connect.hku.hk}

\author{Yupeng Li}
\authornote{Corresponding author.}
\affiliation{%
  \institution{Department of Interactive Media, Hong Kong Baptist University}
  \country{}
  }
\email{ivanypli@gmail.com}

\author{Dacheng Wen}
\affiliation{%
  \institution{Department of Interactive Media, Hong Kong Baptist University}
  \country{}
  }
\affiliation{%
  \institution{School of Computing and Data Science, The University of Hong Kong}
  \country{}
  }
\email{wdacheng@connect.hku.hk}

\author{Yang Chen}
\affiliation{%
  \institution{College of Computer Science and Artificial Intelligence, Fudan University}
  \country{}
  }
\email{chenyang@fudan.edu.cn}

\author{Reynold Cheng}
\affiliation{%
  \institution{School of Computing and Data Science, The University of Hong Kong}
  \country{}
}
\email{ckcheng@cs.hku.hk}

\author{Donglong Chen}
\affiliation{%
  \institution{Beijing Normal-Hong Kong Baptist University}
  \country{}
  }
\email{donglongchen@bnbu.edu.cn}

\author{Francis C. M. Lau}
\affiliation{%
  \institution{School of Computing and Data Science, The University of Hong Kong}
  \country{}}
\affiliation{%
  \institution{Shenzhen Institute of Advanced Technology}
  \country{}}
\email{fcmlau@cs.hku.hk}

\renewcommand{\shortauthors}{Haorui He et al.}

\begin{abstract}
State-of-the-art single-agent claim verification methods struggle with complex claims that require nuanced analysis of multifaceted evidence.
Inspired by real-world professional fact-checkers, 
we propose \textbf{DebateCV}, the first debate-driven claim verification framework powered by multiple LLM agents. 
In DebateCV, two \textit{Debaters} argue opposing stances to surface subtle errors in single-agent assessments.
A decisive \textit{Moderator} is then required to weigh the evidential strength of conflicting arguments to deliver an accurate verdict.
Yet, zero-shot Moderators are biased toward neutral judgments, and no datasets exist for training them.
To bridge this gap, we propose \textbf{Debate-SFT}, a post-training framework that leverages synthetic data to enhance agents' ability to effectively adjudicate debates for claim verification.
Results show that our methods surpass state-of-the-art non-debate approaches in both accuracy (across various evidence conditions) and justification quality.

\end{abstract}



\maketitle

\section{Introduction}
\label{sec:intro}

In the modern digital landscape where misinformation disseminates widely and rapidly online \cite{factcheck_definition}, (automated) claim verification has become a cornerstone web technology for assessing the veracity of (check-worthy) claims \cite{factcheck_definition,guo2022survey,annotation_schema}.
State-of-the-art (SOTA) claim verification approaches typically use retrieval-augmented generation (RAG) \cite{averitec_challenge}, wherein a large language model (LLM) agent receives a claim and relevant evidence retrieved from the web, and generates a verdict with explanatory justifications.

However, the claim verification capabilities of individual agents are insufficient for complex claims that involve multifaceted evidence \cite{conflicts,mohsin2025fundamental}. 
Take \emph{HerO}, the SOTA method using a single agent with supervised fine-tuning (SFT) and Chain-of-Thought (CoT) prompting, as an example.
It incorrectly refutes the claim ``\texttt{52\% of Nigeria's current population lives in urban areas}'' by relying on outdated evidence indicating a 36\% urban population in 2008. \emph{Although more recent evidence supporting a 52\% urbanization rate was successfully retrieved, the single agent in this approach completely overlooked this critical evidence}, resulting in this error.\footnote{This error traces back to Claim 218 in~\citet{hero}'s official results.}

To mitigate errors in individual claim verifications, professional fact-checking agencies in real-world practice commonly adopt a \textbf{debate-driven claim verification} methodology.
For instance, \citet{factcheck_ecology} describes how PolitiFact, a Pulitzer Prize-winning fact-checking agency, employs ``\emph{star chamber}'' sessions, i.e., debates among fact-checkers, to verify claims.\footnote{Please see examples of real-world claim verification debates in \cite{factcheck_ecology}, pp.~204–208.} 
In these sessions, a panel of fact-checkers critically reviews each other's veracity assessments to identify potential flaws in evidence analysis, e.g., the oversight of the latest urbanization statistics in the above example.
\emph{Such an adversarial verification process compels the fact-checkers to iteratively refine their assessments, thereby enhancing the depth and rigour of evidence analysis and fostering holistic claim evaluation} \cite{factcheck_ecology}.
 
Inspired by these real-world debate-driven practices, this work introduces \textbf{DebateCV}, \emph{the first debate-driven claim verification framework based on multiple LLM agents}.\footnote{The initial version of this work was completed in February 2025 \cite{he2025debating}.}
Specifically, DebateCV employs two role-playing \textit{Debater} agents with opposing stances: one affirming and the other refuting the claim's veracity. 
These Debaters iteratively present arguments, contest each other's positions, and defend their own viewpoints by leveraging collected evidence over multiple rounds.
In each round, a third \textit{Moderator} agent evaluates the presented arguments and decides whether further debate is necessary. 
If the debate concludes, the Moderator is tasked with assessing the evidential strength of both sides across all rounds and synthesising an accurate, well-justified verdict.

\begin{figure}[t]
    \centering
    \includegraphics[width=\linewidth]{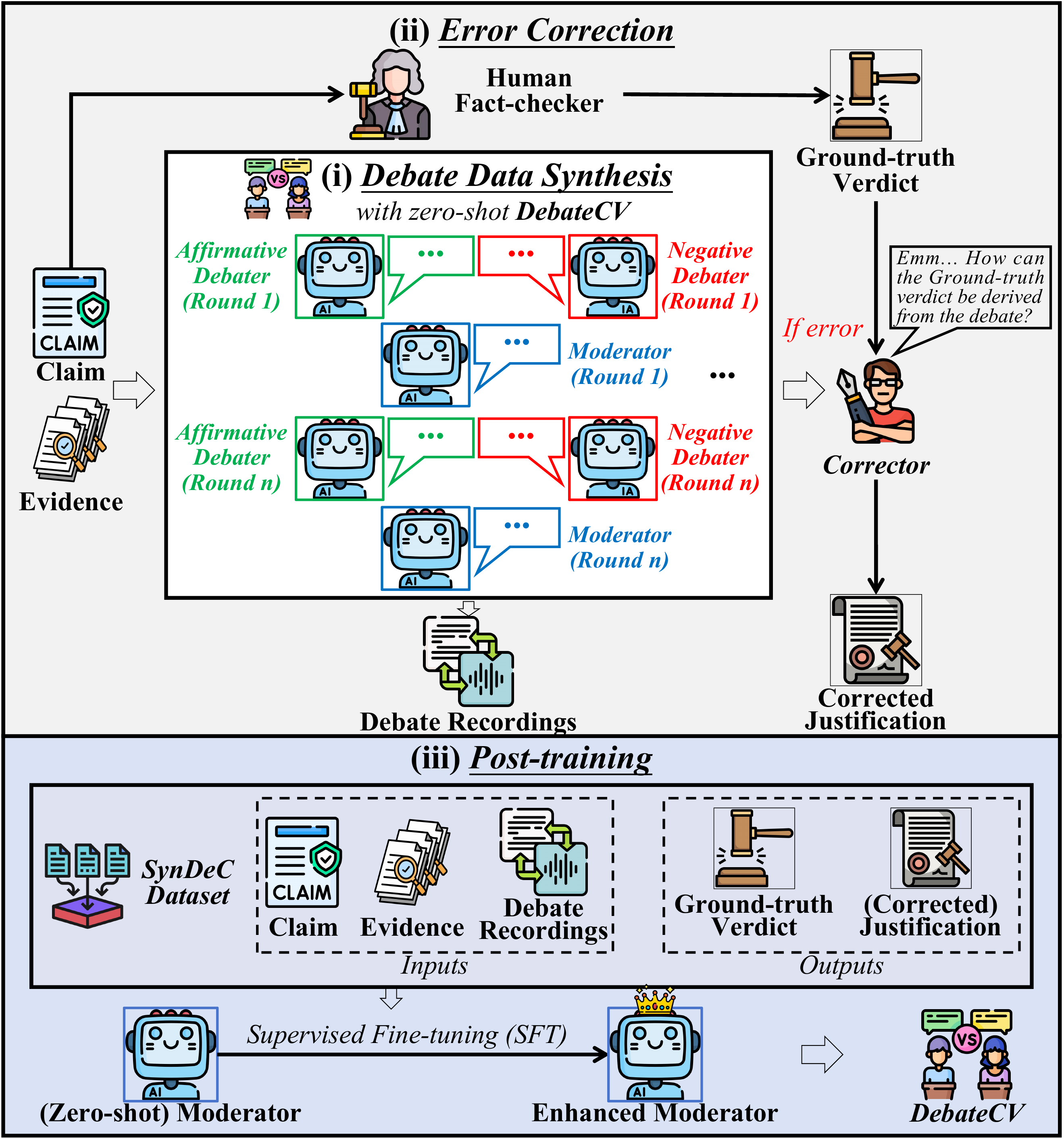}
    \caption{An overview of DebateCV and Debate-SFT.}
    \label{fig:sys}
\end{figure}

\emph{However, our preliminary experiments (see Sec.~\ref{sec:ablation}) show that, while zero-shot LLM agents perform effectively as Debaters, they struggle in the Moderator role}.
For complex claims requiring multiple rounds of debate, they often fail to identify a clear winner, even when sufficient evidence supports a decisive true/false verdict, and instead frequently default to neutral outcomes, such as stating that the evidence is insufficient.
This limitation calls for a dataset containing complete claim verification debate transcripts paired with expert verdicts and justifications grounded in the debates to enable SFT of the Moderator agent. 
Unfortunately, no such dataset exists, and manual construction is prohibitively expensive.
In response, we introduce \textbf{Debate-SFT}, \emph{the first post-training framework specifically designed to fine-tune the Moderator for debate-driven claim verification using synthetic debate data}. 
As illustrated in Fig.~\ref{fig:sys}, our framework begins by constructing the \textbf{SynDeC} (\underline{Syn}thetic \underline{De}bate for \underline{C}laim verification) dataset, 
which is generated by executing zero-shot DebateCV to synthesise multi-round debate transcripts for various claims. 
We then identify and correct discrepancies between the zero-shot Moderator's outputs and ground-truth labels to ensure a clean training signal. 
The Moderator is subsequently post-trained on the curated SynDeC dataset to enhance nuanced argument evaluation. 
With the enhancement, DebateCV leverages a ``star chamber'' among LLM agents to accurately verify claims.

\textbf{Experiments} show that DebateCV outperforms existing SOTA methods by 2.6–5.8\% in accuracy across diverse evidence quality settings.
\textbf{Case studies} illustrate that DebateCV's adversarial debate mechanism effectively corrects typical single-agent errors, such as claim/evidence misinterpretation, overlooked evidence, and overreliance on speculation. 
\textbf{Human expert evaluations} further confirm that DebateCV provides higher-quality justifications, beyond accuracy improvements.
Our contributions are as follows.\footnote{Our code is available at \url{github.com/TrustworthyComp/DebatingTruth}}

\begin{itemize}[leftmargin=*]
\item \textbf{Verification Framework.} We propose \textbf{DebateCV}, the first debate-driven claim verification method with multiple LLM agents.
\item \textbf{Post-training Framework.} We propose \textbf{Debate-SFT}, a novel approach that generates the \textbf{SynDeC} dataset to fine-tune LLMs for robust adjudication in claim verification debates.
\item \textbf{Evaluation.} We conduct experiments to show that our methods surpass existing SOTAs in both accuracy and justification quality.
\end{itemize}

\section{Related Work}
\paragraph{\textbf{Claim Verification.}}
Automated fact-checking systems \cite{infact,hero,competing_wisdom,guo2022survey,leippold2025automated,CTU_at_averitec} typically operate in two stages: first, they retrieve relevant evidence (\emph{evidence retrieval} stage); then, they use a single LLM agent to predict a verdict, such as \textit{Supported}, \textit{Refuted}, \textit{Not Enough Evidence}, or \textit{Conflicting Evidence/Cherry-picking}, along with a justification \cite{averitec_challenge,hero} (\emph{claim verification} stage).\footnote{Appendix~\ref{sec:definition} provides the definition of the four verdict categories.}

However, as shown in Sec.~\ref{sec:intro}, existing single-agent systems are vulnerable to multifaceted and even misleading evidence retrieved from the open web. 
This vulnerability can arise from a fundamental mismatch: LLMs are pretrained to model linguistic patterns by autoregressively predicting the next token, while claim verification demands comprehensive analysis of collected evidence~\cite{mohsin2025fundamental}. 
Ensembling multiple agents, e.g., via majority voting~\cite{debate_or_vote, fact_audit} or collaborative frameworks~\cite{camel}, 
also lacks rigorous adversarial scrutiny and may amplify individual errors instead of correcting them.
Inspired by real-world fact-checking practices \cite{factcheck_ecology}, 
we propose the first debate-driven claim verification framework, DebateCV, to expose and resolve individual assessment errors.

Close to our work, 
a few recent studies \cite{competing_wisdom, yue2024retrieval} have explored the value of integrating evidence or arguments from opposing perspectives. 
\citet{competing_wisdom} retrieve and present separate sets of supporting and opposing evidence to a single LLM for verdict prediction, 
which focuses on improving \emph{evidence retrieval}. 
Instead, we target the subsequent \emph{claim verification} stage by enabling comprehensive analysis of retrieved evidence. 
\citet{yue2024retrieval} generate static supporting and refuting arguments independently, 
but their approach lacks an interactive debate mechanism to surface flaws or resolve contradictions, a mechanism that DebateCV mimics from human fact-checking. 
Another concurrent work \cite{han2025beyond} investigates evidence-based multi-agent debate for misinformation detection, 
but it is limited to binary (true/false) classification. 
This oversimplifies real-world scenarios where claims may lack sufficient evidence (\textit{Not Enough Evidence}) or involve ambiguity and source disagreement (\textit{Conflicting Evidence/Cherry-picking}) \cite{averitec}. 
Also, they rely solely on zero-shot LLMs, 
which, as shown in Sec.~\ref{sec:ablation}, 
exhibit conformity bias and frequently default to neutral verdicts in realistic non-binary settings.
In contrast, we introduce Debate-SFT, the first post-training framework that substantially improves LLMs' ability to adjudicate complex debates.
We are also the first to conduct a human expert evaluation of justification quality (see Sec.~\ref{sec:main}) to demonstrate the superior transparency of DebateCV framework.

\paragraph{\textbf{Multi-agent Debate.}}
Multi-agent interactions have empowered diverse applications, 
such as AI societies \cite{camel} and software engineering \cite{multiagent_survey, metagpt, autogen},
with multi-agent debate-based frameworks often outperforming other forms in complex tasks, such as machine translation \cite{mad},  hallucination mitigation \cite{debate_improve_factuality, debaterag}, and stance detection \cite{lan2024stance}. 
However, prior approaches to misinformation detection \cite{debate_for_fake, debate_to_detect, liu2025truth} mainly debate linguistic patterns of fake news without retrieving external evidence.
These patterns are typically platform-specific \cite{www_mcfend} or time-constrained \cite{yang2024search}, 
limiting their real-world applicability.
In contrast, this work targets the claim verification task that analyses consistencies and contradictions between retrieved evidence and claims \cite{guo2022survey}, 
and is the first to propose a debate-driven framework for this task.

Beyond downstream applications, three prior studies have explored leveraging debate mechanisms for post-training LLMs. 
\citet{debategpt} employ debates to refine training corpus annotations for general language modelling. 
\citet{debat_for_diverse} use debates to generate convincing arguments for fine-tuning LLM persuasiveness. 
\citet{acc_dabte} use debate games between two LLM teams to improve their team-working abilities.
Different from these works, we propose Debate-SFT, the first post-training framework specifically designed for enhancing LLMs' ability to distil multi-round debates into accurate verdicts.
We note that, though tailored for claim verification, Debate-SFT, with minimal modifications, can be adapted to other contexts using multi-agent debate.

\section{DebateCV: Debate-driven Claim Verification}\label{sec:debatecv}
 
Debate-driven claim verification leverages multi-round debates for thorough evidence analysis. 
We formalize this task as determining the veracity of a textual claim \(c\) using an evidence set \(\mathcal{E} = \{e_1, \ldots, e_k\}\) by simulating a debate between specialised LLM agents.
The outcome is a verdict \(\hat{y}\) and a corresponding justification \(\hat{j}\), both grounded in the arguments exchanged during the debate.
In this section, we present \textbf{DebateCV} and describe how LLM agents are coordinated within the framework.

\subsection{Pre-debate Configuration}
Similar to debates conducted by humans, debates conducted by LLM agents must be carefully configured through role assignment. 
DebateCV employs three distinct LLM agents with complementary responsibilities: two Debaters, namely an Affirmative Debater \(D^+\) and a Negative Debater \(D^-\), and one Moderator \(M\). 
Each agent is initialized with tailored system prompts that define their behaviour throughout the debate. Their roles are outlined as follows.

\paragraph{\textbf{Debaters.}}
The Affirmative Debater \(D^+\) and Negative Debater \(D^-\) are tasked with examining the veracity of the given claim \(c\) from opposing perspectives. \(D^+\) advocates for the claim's validity, while \(D^-\) challenges its truthfulness. Both agents have access to the identical evidence set \(\mathcal{E} = \{e_1, e_2, \ldots, e_k\}\), ensuring arguments are based on a shared factual foundation. Each Debater constructs evidence-based arguments, defends their assigned stance, and critically counters the opponent's points using the available evidence.

\paragraph{\textbf{Moderator.}}
The Moderator \(M\) is expected to act as an impartial arbiter overseeing the debate. Its primary responsibilities include synthesising concise round summaries \(S_t\), detecting convergence to enable early termination when no new insights are presented, and determining the final verdict \(\hat{y}\) for claim \(c\) with a justification \(\hat{j}\) that distils the key insights from the debate.

\subsection{In-Debate Process}
 
After pre-debate configuration, the agents engage in a structured, multi-round debate with a strict turn order. In the first round ($t = 1$), the Affirmative Debater ($D^+$) presents an opening argument ($A^+_1$), citing the most relevant supporting evidence from $\mathcal{E}$ to establish the validity of the claim $c$. The Negative Debater ($D^-$) then produces a rebuttal ($A^-_1$) that challenges this opening argument, also drawing on evidence from $\mathcal{E}$.
In subsequent rounds ($t > 1$), the Affirmative Debater ($D^+$) responds with a counterargument ($A^+_t$) defending against the previous rebuttal ($A^-_{t-1}$). This exchange continues, alternating between rebuttals and responses, for each round.

At the end of each round, the Moderator ($M$) synthesises the key arguments and evidence from both sides into a summary of all exchanged arguments ($S_t$). The Moderator then assesses whether the debate has converged, e.g., it deems that $S_t$ and $S_{t-1}$ are highly similar, indicating no new insights are being introduced. The debate terminates early if $M$ detects convergence; otherwise, it proceeds until the maximum number of rounds ($T_{\text{max}}$) is reached. Upon conclusion, $M$ determines the final verdict ($\hat{y}$) by weighing the arguments from both sides and provides a justification ($\hat{j}$).

The specific prompts used are provided in Appendix~\ref{sec:prompt}.

\section{Debate-SFT: Post-training the Moderator}\label{sec:synthetic}
The DebateCV framework, introduced in Sec.~\ref{sec:debatecv}, provides a debate-driven approach to claim verification. 
In our preliminary experiments, we find that zero-shot Debater agents consistently produce well-reasoned arguments from their assigned perspectives (see Sec.~\ref{sec:validation} for human evaluation results regarding debate quality). However, the zero-shot \textit{Moderator} often struggles with complex claims that require multiple debate rounds and tends to exhibit conformity bias~\cite{conform1,conform2}. Specifically, it frequently defaults to neutral verdicts such as \textit{Not Enough Evidence} or \textit{Conflicting Evidence/Cherry-picking}, even when sufficient evidence exists to support a definitive judgement (see Sec.~\ref{sec:ablation} for experimental results).

To enhance the LLM agents' ability to adjudicate debates, we aim to employ supervised fine-tuning (SFT), a well-established approach for adapting LLMs to downstream tasks~\cite{post-train_survey}. 
However, SFT requires annotated debate-driven claim verification datasets,
which contain complete multi-round debate transcripts paired with final verdicts and justifications grounded in the debates. 
Such datasets are both scarce and expensive to curate.
To address this challenge, we introduce \textbf{Debate-SFT}, a novel post-training framework that transforms existing non-debate claim verification datasets into synthetic debate-driven equivalents, enabling effective fine-tuning of the Moderator. This section outlines the three stages of Debate-SFT: (1)~\textit{debate data synthesis}, (2)~\textit{error correction}, and (3)~\textit{post-training}.

\paragraph{\textbf{Debate Data Synthesis.}}
Given a non-debate claim verification dataset with \( n \) claims \( \mathcal{C}=\{c_1, c_2, \dots, c_n\} \), their ground-truth verdicts \( \mathcal{Y}=\{y_1, y_2, \dots, y_n\} \), and golden evidence sets \( \mathcal{E}=\{\mathcal{E}_1, \mathcal{E}_2, \dots, \mathcal{E}_n\} \) annotated by humans, we leverage the zero-shot DebateCV framework to generate synthetic debate recordings \( \mathcal{D}=\{d_1, d_2, \dots, d_n\} \). Each recording \( d_i \) is a multi-round transcript of the arguments and summaries produced during the debate over claim \( c_i \):  
$
d_i = \bigl\{ A^+_1, A^-_1, \dots, A^+_{t_i}, A^-_{t_i} \bigr\},
$  
where \( A^+_{t_i} \) and \( A^-_{t_i} \) are the affirmative and negative arguments in round \( {t_i} \).\footnote{Our human evaluation in Sec.~\ref{sec:validation} shows that the generated debates feature well-reasoned, evidence-based arguments and consistently uphold opposing stances.}
 During this process, the zero-shot Moderator predicts verdicts \( \mathcal{\hat{Y}}=\{\hat{y}_1, \hat{y}_2, \dots, \hat{y}_n\} \) with justifications \( \mathcal{\hat{J}}=\{\hat{j}_1, \hat{j}_2, \dots, \hat{j}_n\} \) based on $\mathcal{D}$.

\paragraph{\textbf{Error Correction.}}
Since the zero-shot Moderator may issue erroneous verdicts even when the debaters' arguments are established in gold-standard evidence, we use human-rated ground-truth verdicts to ensure data validity. 
Specifically, when the predicted verdict \( \hat{y}_i \) differs from the ground-truth \( y_i \), we introduce an LLM-based Corrector to generate a justification that aligns with the human annotation. For each claim \( c_i \) where \( \hat{y}_i \neq y_i \), the Corrector takes the full debate recording \( d_i \) and the human-rated ground-truth verdict \( y_i \) as input to generate a corrected justification \( j'_i \) that explains how \( y_i \) can be derived from the arguments and evidence in \( d_i \).\footnote{Our human evaluation in Sec.~\ref{sec:validation} shows that these Corrector-generated justifications effectively reflect the debated content and align with human judgements on the verdict.}

\paragraph{\textbf{Post-training.}}\label{sec:post-train}
With the data synthesised in the previous two stages, we can now conduct supervised fine-tuning to enhance the Moderator's performance in debate-driven claim verification. Specifically, we divide the claims into two groups based on the zero-shot Moderator's verdict accuracy. For \textbf{correct samples} (\( \mathcal{C}_{\text{correct}} = \{ c_i \mid \hat{y}_i = y_i \} \)), we use the debate recording \( d_i \) as multi-turn dialogue context and fine-tune the Moderator to generate the correct verdict \( \hat{y}_i \) and its corresponding justification \( \hat{j}_i \). For \textbf{error samples} (\( \mathcal{C}_{\text{error}} = \{ c_i \mid \hat{y}_i \neq y_i \} \)), we use the debate recording \( d_i \) as context and fine-tune the Moderator to generate the ground-truth verdict \( y_i \) from human annotation and the corrected justification \( j'_i \). By combining both types of samples in training, the Debate-SFT framework enables the Moderator to learn from examples where it was originally correct and from cases where errors were corrected:
\[
\theta_{\text{Debate-SFT}} = \arg\min_{\theta} \mathcal{L}_{\text{SFT}}(\mathcal{C}_{\text{correct}} \cup \mathcal{C}_{\text{error}}, \theta),
\]
where \( \mathcal{L}_{\text{SFT}} \) is the standard SFT loss \cite{sft} and \( \theta \) represents the zero-shot Moderator parameters. This three-stage Debate-SFT framework results in an enhanced Moderator that can better evaluate debate dynamics and render more accurate judgements.

\paragraph{\textbf{Resulting SynDeC Dataset.}}
To instantiate the Debate-SFT framework,
we apply it to the training split of AVeriTeC~\cite{averitec} (3,068 claims) to create the first synthetic dataset designed for debate-driven claim verification, \textbf{SynDeC}.\footnote{Our method can be applied to any non-debate claim verification dataset with human-rated verdicts and evidence.}  
Specifically, we use the zero-shot DebateCV framework to synthesise full debate recordings \(d_i\) for all 3,068 claims. 
Among these, 936 claims are incorrectly predicted and undergo error correction, where a Corrector based on GPT-4o generates a corrected justification \(j'_i\) aligned with the ground-truth verdict \(y_i\).  
Each sample in the SynDeC dataset contains the following elements:
(i) claim \(c_i\),  
(ii) evidence set \(\mathcal{E}_i\),  
(iii) debate recording \(d_i\),
(iv) ground-truth verdict \(y_i\),  
(v) predicted verdict \(\hat{y}_i\),  
(vi) predicted justification \(\hat{j}_i\),  
and, when \(\hat{y}_i \neq y_i\), (vii) the corrected justification \(j'_i\).

\section{Experiments}\label{sec:experiment}
In this section, we conduct experiments to evaluate our methods. 
\subsection{Experimental Setup}
\paragraph{\textbf{Dataset and Evidence Conditions.}}
To evaluate DebateCV, we use AVeriTeC \cite{averitec}, the state-of-the-art (SOTA) real-world claim verification benchmark, which addresses the evidence leakage and insufficiency issues in prior datasets \cite{defame,fact2fiction} like FEVER \cite{fever}.
AVeriTeC consists of English textual claims collected from 50 fact-checking agencies and annotated by human fact-checkers.
Each claim in AVeriTeC is paired with a knowledge base containing both relevant and potentially distracting evidence collected via Google Search, which simulates open-web retrieval while ensuring reproducibility \cite{averitec_challenge}.
To evaluate our method under different evidence quality conditions, we consider three distinct evidence settings:
\begin{itemize}[leftmargin=*]
    \item \textbf{Golden}: Human-collected evidence that represents the ideal scenario with perfect evidence retrieval. This condition provides an upper bound for claim verification performance.
    \item \textbf{Retrieved (H)}: Evidence retrieved using the HerO framework \cite{hero}, which achieved the best evidence retrieval performance in the AVeriTeC challenge \cite{averitec_challenge}. This condition simulates realistic end-to-end fact-checking pipelines where retrieval may introduce irrelevant or misleading evidence.
    We use the pre-retrieved evidence provided by \citet{hero}.\footnote{\url{https://github.com/ssu-humane/HerO/}}
    \item \textbf{Retrieved (I)}: Evidence retrieved using the InFact framework \cite{infact}, which ranked second in the challenge above.
    This additional condition allows us to assess the robustness of DebateCV across different evidence retrieval strategies. We use the official InFact implementation with GPT-4o-Mini-2024-07-18 as the backbone.\footnote{\url{https://github.com/multimodal-ai-lab/DEFAME/tree/infact}}
\end{itemize}

For training, we use the train split of the AVeriTeC dataset (3,068 claims).
For evaluation, we use its development split (500 claims), which remains unseen during training to mitigate bias.\footnote{We exclude the AVeriTeC test split because its golden evidence and ground-truth labels are not publicly available.}

\paragraph{\textbf{Baselines and Ablations.}}
We compare DebateCV against SOTA \textbf{single-agent} and \textbf{multi-agent} baselines, and \textbf{ablation variants}.

\textbf{Single-agent} baselines prompt a single LLM to use the provided evidence to predict the verdict of the claims. 
SOTA methods employ the chain-of-thought (CoT) prompting strategy \cite{cot} to enhance the reasoning capabilities of zero-shot LLMs. 
Specifically, we evaluate two such prompting strategies: \textit{CoT} \cite{yang_and_rocha}, and \textit{InFact} \cite{infact}.
In addition to zero-shot methods, we also include (3) \textit{HerO} \cite{hero}, which conducts supervised fine-tuning (SFT) on the training split of the AVeriTeC dataset to further improve the performance of single LLMs.

\textbf{Multi-agent} baselines include:  
(1)~\textit{Majority}: \citet{fact_audit} propose employing three independent agents to verify each claim, with the final verdict determined by majority voting among the agents. In our implementation, each agent uses the CoT baseline, and we adopt the majority voting prompt from the official implementation of \citet{fact_audit}.\footnote{\url{https://github.com/DanielLin97/FACT-AUDIT}}
(2)~\textit{Cooperation}: This baseline utilises the CAMEL framework \cite{camel}, which simulates collaborative task-solving through structured interactions between an AI user and an AI assistant. We leverage the official implementation of CAMEL and use the prompt of the CoT baseline as the task description.\footnote{\url{https://github.com/camel-ai/camel}}

For \textbf{ablation variants} of DebateCV, we include:  
(1)~\textit{w/o \( \mathcal{C}_{\text{correct}} \)}: the Moderator is trained exclusively on corrected error samples. This variant evaluates the effectiveness of our error correction mechanism.  
(2)~\textit{w/o \( \mathcal{C}_{\text{error}} \)}: the Moderator is trained only on originally correct samples (\( \mathcal{C}_{\text{correct}} \)). This variant assesses the contribution of the originally correct samples.  
(3)~\textit{w/o Debate-SFT}: the Moderator is not post-trained. This variant evaluates the importance of our Debate-SFT method and the constructed SynDeC dataset.

\paragraph{\textbf{Implementation Details.}}
We evaluate the baselines using four LLMs: two proprietary models, GPT-4o-2024-05-13 (GPT-4o) and GPT-4o-Mini-2024-07-18 (GPT-4o-Mini), and two open-source models, Llama-3.1-8B \cite{llama} and Qwen-2.5-7B \cite{qwen}.  
For HerO, which requires supervised fine-tuning, we restrict our experiments to the open-source models due to the unavailability of trainable parameters for GPT-4o and GPT-4o-Mini.  
In DebateCV, the Debaters employ GPT-4o-Mini to balance cost efficiency with strong performance.\footnote{Sec.~\ref{sec:llm_impact} validates that DebateCV is robust to alternative Debaters.}
The Moderator, which produces the final verdict and justification, uses Llama-3.1-8B and Qwen-2.5-7B to enable a direct comparison with HerO \cite{hero}.
This configuration ensures a comprehensive evaluation across diverse LLM architectures.

We set the maximum number of generated tokens to 512, temperature to 0.7, and top\_p to 1.0.  
For post-training, we employ LoRA with rank 128 and alpha 256 and the Adam optimizer with a learning rate of 2e-5 for 2 epochs.  
These hyper-parameters are identical to those used by \citet{hero} to ensure a fair comparison.

\paragraph{\textbf{Evaluation Metrics.}} 
We employ two complementary metrics to evaluate the methods. 
(1) \textbf{Accuracy (Acc.)}: The proportion of correctly classified claims across all test instances, which measures an overall performance regardless of evidence sufficiency.
(2) \textbf{AVeriTeC Score (Aver.)}: Accuracy conditioned on evidence sufficiency \cite{averitec}, which excludes correctly classified claims that lack sufficient supporting evidence, thereby avoiding score inflation due to lucky guesses. 
We follow \citet{averitec} to use the METEOR score \cite{meteor} to assess evidence sufficiency between retrieved evidence $\hat{\mathcal{E}}$ and golden evidence $\mathcal{E}$, denoted as $f(\hat{\mathcal{E}}, \mathcal{E})$.
Claims are deemed to have sufficient evidence if  $f(\hat{\mathcal{E}}, \mathcal{E}) \geq 0.25$ \cite{averitec}.\footnote{Note that under the golden evidence condition, the AVeriTeC score equals accuracy; therefore, we report only accuracy for this setting.}
For both metrics, we follow \citet{complex_method} to use paired bootstrap tests to assess statistical significance based on at least five repeated trials. Results with $p \leq 0.05$ compared to the strongest baseline (HerO) are marked with ``$\dagger$''.

\subsection{Comparison with Existing Methods}\label{sec:main}

\paragraph{\textbf{Main Results.}}
\begin{table}[h!]
\centering
\caption{Claim verification performance across evidence conditions (reported as percentages). The best result in each column is bolded; the second-best is underlined.}
\label{tab:main}
\resizebox{\columnwidth}{!}{%
\begin{tabular}{lllllll}
\toprule
\multirow{2}{*}{\textbf{Methods}} & \multirow{2}{*}{\textbf{LLMs}} &
\multicolumn{1}{c}{\textbf{Golden}} &
\multicolumn{2}{c}{\textbf{Retrieved (H)}} &
\multicolumn{2}{c}{\textbf{Retrieved (I)}} \\
\cmidrule(lr){3-3} \cmidrule(lr){4-5} \cmidrule(lr){6-7}
& & Acc.  & Acc. &AVer. & Acc. &AVer. \\
\midrule
\multirow{4}{*}{CoT} 
& GPT-4o & 81.6 & 63.8 & 47.8 & 66.6 & 51.4 \\
& GPT-4o-Mini & 80.8 & 62.8 & 46.8 & 65.8 & 51.0 \\
& Llama-3.1-8B & 76.8 & 62.4 & 46.6 & 65.0 & 50.2 \\
& Qwen-2.5-7B & 77.0 & 62.6 & 46.8 & 65.4 & 50.6 \\
\midrule
\multirow{4}{*}{InFact} 
& GPT-4o & 81.2 & 68.6 & 52.6 & 68.8 & 52.6 \\
& GPT-4o-Mini & 79.2 & 65.2 & 47.8 & 67.4 & 51.4 \\
& Llama-3.1-8B & 73.2 & 63.4 & 47.0 & 65.2 & 50.4 \\
& Qwen-2.5-7B & 74.4 & 63.6 & 47.4 & 65.4 & 50.6 \\
\midrule
\multirow{2}{*}{HerO} 
& Llama-3.1-8B & 80.4 & 70.2 & 52.6 & 67.8 & 51.8 \\
& Qwen-2.5-7B & 80.0 & 68.2 & 50.6 & 66.8 & 51.2 \\
\midrule
\multirow{4}{*}{Majority} 
& GPT-4o & 81.8 & 64.4 & 49.2 & 67.2 & 51.6 \\
& GPT-4o-Mini & 81.4 & 63.8 & 47.8 & 66.2 & 51.4 \\
& Llama-3.1-8B & 78.0 & 63.8 & 47.6 & 65.2 & 49.4 \\
& Qwen-2.5-7B & 77.6 & 63.6 & 48.2 & 65.4 & 49.8 \\
\midrule
\multirow{4}{*}{Cooperation} 
& GPT-4o & 81.8 & 67.6 & 51.2 & 68.4 & 52.2  \\
& GPT-4o-Mini & 81.6 & 65.2 & 49.4 & 66.2 & 51.6 \\
& Llama-3.1-8B & 76.6 & 61.2 & 46.8 & 64.8  & 51.0 \\
& Qwen-2.5-7B & 77.4 & 60.8 & 46.4 & 64.4 & 50.6 \\
\midrule
\multirow{2}{*}{DebateCV} 
& Llama-3.1-8B & \textbf{83.4}$^\dagger$ &  \textbf{72.8}$^\dagger$ &  \textbf{54.4}$^\dagger$ &  \textbf{73.6}$^\dagger$ &  \textbf{54.8}$^\dagger$ \\
& Qwen-2.5-7B & \underline{82.0}$^\dagger$ & \underline{72.6}$^\dagger$ & \underline{54.6}$^\dagger$ & \underline{72.4}$^\dagger$ & \underline{54.6}$^\dagger$ \\
\midrule
\multirow{2}{*}{\quad w/o \( \mathcal{C}_{\text{error}} \)} 
& Llama-3.1-8B & 76.4 & 63.0 & 49.2 & 68.2 & 51.8 \\
& Qwen-2.5-7B & 75.8 & 62.9 & 49.1 & 66.4 & 51.2 \\
\midrule
\multirow{2}{*}{\quad w/o \( \mathcal{C}_{\text{correct}} \)} 
& Llama-3.1-8B & 76.8 & 70.8 & 53.8 & 71.6 & 54.2 \\
& Qwen-2.5-7B & 75.6  & 69.4 & 53.0 & 71.2 & 53.8 \\
\midrule
\multirow{4}{*}{\quad w/o Debate-SFT} 
& GPT-4o & 78.8 & 65.4 & 48.6 & 66.2 & 50.2 \\
& GPT-4o-Mini & 74.6 & 54.8 & 40.4 & 61.2 & 46.8 \\
& Llama-3.1-8B & 66.6 & 53.8 & 41.2 & 56.2 & 43.2 \\
& Qwen-2.5-7B & 61.4 & 54.2 & 41.0 & 56.0 & 42.8 \\
\bottomrule
\end{tabular}
}
\end{table}

Table~\ref{tab:main} compares DebateCV against state-of-the-art baselines across three evidence conditions.

Zero-shot single-agent baselines (CoT and InFact) and multi-agent approaches (Majority and Cooperation) both exhibit limited robustness under realistic retrieved evidence settings.
For example, when using GPT-4o as the backbone model, CoT's accuracy drops from 81.6\% with golden evidence to 63.8\% in the Retrieved (H) setting by 17.8\%.
Multi-agent methods leverage collective intelligence but yield only modest improvements over single-agent counterparts; for instance, Majority achieves merely a 0.6 point gain over CoT with GPT-4o under both Retrieved (H) and Retrieved (I) conditions, while Cooperation, which simulates user-assistant dialogue for evidence analysis, only improves with strong LLMs like GPT-4o but can perform worse with smaller models, such as Qwen-2.5-7B in Retrieved (H), where accuracy drops from 62.6\% to 60.8\%.
The results underscore the critical challenge of handling noisy retrieved evidence in end-to-end fact-checking pipelines.
\textbf{These limitations stem from the absence of cross-validation mechanisms in existing approaches.}
Single-agent reasoning follows a linear chain of thought, rendering it vulnerable to misleading or incomplete evidence (for example, those introduced in Sec.~\ref{sec:intro} and \ref{sec:case}).
Multi-agent baselines also lack rigorous adversarial critique, reinforcing erroneous consensus rather than correcting it.

Our debate-driven claim verification framework, \textbf{DebateCV}, consistently outperforms all baselines across every evidence condition. For instance, when powered by Llama-3.1-8B, DebateCV achieves the highest scores: \textbf{83.4\%} under the golden evidence setting, \textbf{72.8\% / 54.4\%} under Retrieved (H), and \textbf{73.6\% / 54.8\%} under Retrieved (I).  
These results demonstrate that \textbf{our debate-driven approach significantly surpasses the prior state-of-the-art HerO method}~\cite{hero}, delivering absolute gains of \textbf{at least 2.0\%} under ideal golden evidence and \textbf{2.6--5.8\%} under realistic retrieved evidence conditions. This superiority arises from the adversarial cross-agent critique inherent in the debate process, which rigorously evaluates evidence and effectively mitigates the influence of misleading or ambiguous information (see the qualitative case study in Sec.~\ref{sec:case} for empirical support for this finding).
The DebateCV variant using Qwen-2.5-7B as its backbone also delivers strong performance, securing the \textbf{second-best results} across all settings: 82.0\% (Golden), 72.6\% / 54.6\% (Retrieved (H)), and 72.4\% / 54.6\% (Retrieved (I)), respectively. This underscores the \textbf{robustness of our framework across different LLM backbones}.

\paragraph{\textbf{Human Evaluation on Justification Quality}}

\begin{table}[h!]
\centering
\caption{Mean Opinion Score results on justification quality.}
\label{tab:justification_quality}
\resizebox{\linewidth}{!}{%
\begin{tabular}{lccc}
\toprule
\textbf{Criteria/Methods} & \textbf{HerO} & \textbf{DebateCV} & \textbf{Win / Tie / Loss} \\
\midrule
Evidence Used & 2.60 & \textbf{3.67}$^\dagger$ & 57.6\% / 27.5\% / 14.9\%  \\
Sources of Uncertainty & 2.22 & \textbf{3.07}$^\dagger$ & 50.5\% / 32.8\% / 16.7\%  \\
Reasoning Pathway & 2.49 & \textbf{3.59}$^\dagger$ & 55.6\% / 28.3\% / 16.1\%  \\
\bottomrule
\end{tabular}
}
\end{table}
Beyond accuracy, we conduct a Mean Opinion Score (MOS) evaluation \cite{mos} of justification quality based on the criteria proposed by \citet{warren2025show}, who identified three key dimensions of justification quality by interviews with fact-checking experts across five continents:
(1) \textbf{Evidence Used}: clarity and comprehensiveness in presenting supporting evidence;
(2) \textbf{Sources of Uncertainty}: appropriate acknowledgment of evidence limitations; and
(3) \textbf{Reasoning Pathway}: explicit, logical connections from evidence to conclusion.

For this evaluation,
we randomly sampled 50 claims from our evaluation set that were correctly verified by both DebateCV and the best-performing baseline, HerO, under the Retrieved (H) condition.
We recruited 36 participants, each with at least postgraduate-level expertise in communications or journalism and proficient in English, to rate the justifications.
Each participant was given the claim and retrieved evidence, and scored the justifications from both methods (presented in randomized order) on all three dimensions using a 5-point scale (1 = very poor, 5 = very good).
The rating for each justification is averaged across at least three participants.

As shown in Table~\ref{tab:justification_quality}, DebateCV significantly outperforms HerO across all three dimensions, with MOS gains of 1.07, 0.85, and 1.10 points, respectively.
In head-to-head preference, human experts chose DebateCV's justification as superior in 50.5--57.6\% of cases. 
This demonstrates that DebateCV produces markedly more trustworthy and expert-aligned justifications than existing SOTA.

\paragraph{\textbf{Case Study.}} \label{sec:case}
\begin{table*}[thbp!]
\centering
\small
\caption{The claim verification results of DebateCV. Responses are condensed for brevity while preserving original semantics. Erroneous judgements are in \flaw{purple}, accurate judgements are in \good{green}, and adversarial arguments in DebateCV are in \adver{blue}.}
\resizebox{\textwidth}{!}{%
\begin{tabular}
{p{0.18\textwidth}p{0.20\textwidth}p{0.21\textwidth}p{0.19\textwidth}p{0.19\textwidth}p{0.20\textwidth}}
\midrule
\multirow{2}{*}{\centering \textbf{Claim (ID; Label)}} &
\multirow{2}{*}{\centering \textbf{CoT}} &
\multirow{2}{*}{\centering \textbf{HerO}} &
\multicolumn{3}{c}{\textbf{DebateCV}} \\
\cline{4-6}
& & &\textbf{Affirmative Debater} & \textbf{Negative Debater} & \textbf{Moderator} \\
\midrule
Amy Coney Barrett was confirmed as US Supreme Court Justice on Oct.~26, 2020. 
\textbf{(ID: 31; \textit{Supported})}
&
\flaw{Most evidence supports Oct.~26, but some also say Oct.~27.} This leads to Conflicting Evidence. \flaw{\XSolidBrush}
&
\good{Multiple evidences confirm the Senate voted 52-48 to confirm Barrett on October 26, matching the claim date.} \textit{Supported}. \textcolor{green}{\Checkmark}
&
Multiple sources state she was confirmed as a Supreme Court Justice on Oct.~26 via a 52–48 vote. \textit{Supported}.
&
The term ``confirmed'' \adver{can be interpreted as} the date she officially assumed duty, which is Oct.~27. \textit{Refuted}.
&
\good{The claim refers to the confirmation of her role, not the official swearing-in.} The claim is \textit{Supported}. \textcolor{green}{\Checkmark} \\ 
\midrule
New Zealand's Abortion Act does not mandate medical support for babies born alive after an abortion.  
\textbf{(ID: 99; \textit{Refuted})}
&
Multiple evidence confirms that \flaw{an amendment to the Act proposing this duty was rejected.} The claim is \textit{Supported}. \flaw{\XSolidBrush}
&
\flaw{Evidence shows there is no requirement for a baby born alive after an abortion to receive medical support.} \textit{Supported}. \flaw{\XSolidBrush}
&
Multiple evidence confirm there is no such legal duty; the rejection of the amendments is also a supporting evidence. \textit{Supported}.
&
The Act \adver{clearly states that} "... have a duty to provide medical care." The amendments \adver{were rejected as redundant.} \textit{Refuted}.
&
The Negative argument is \good{more strongly supported by the Act itself}, and the rejection of the amendments further reinforces this position. \textit{Refuted}. \textcolor{green}{\Checkmark}\\
\midrule
Premier Daniel Andrews of Victoria in Australia sold the rights to water to China.  
\textbf{(ID: 282; \textit{Not Enough Evidence})}
&
Evidence shows the premier does not intend to change his approach to China, \flaw{which implies no current plans to sell water rights}. \textit{Refuted}. \flaw{\XSolidBrush}
&
\flaw{Evidence shows Andrews signed trade agreements and China owns water entitlements, but the water was not sold to China.} \textit{Refuted}. \flaw{\XSolidBrush}
&
The Premier fostered an environment conducive to foreign ownership of water rights; China holds significant water entitlements. \textit{Supported}.
&
The Premier promoted business cooperation, \adver{not water rights deals}. China's ownership \adver{does not equate to} approved sales. \textit{Refuted}.
&
Arguments from \good{both sides rely more on interpretation and inference than on concrete evidence of a transaction.} \textit{Not Enough Evidence}. \textcolor{green}{\Checkmark}
\\
\bottomrule
\end{tabular}
\label{tab:case_study}
}
\end{table*}

To interpret the performance gain in accuracy and justification quality, we conduct a case study to compare the claim verification results of DebateCV with those of the CoT and HerO. 
We select three representative claims from the evaluation set, each corresponding to a different verdict category: \textit{Supported} (Claim 31), \textit{Refuted} (Claim 99), and \textit{Not Enough Evidence} (Claim 282).\footnote{Ten additional examples are provided in Appendix~\ref{sec:examples}.} 
As shown in Table~\ref{tab:case_study}, \textbf{the baselines exhibit three error patterns that are addressed by our adversarial debate mechanism}.

\textbf{(1) Claim/evidence misinterpretation}. 
For Claim 31 on Amy Barrett's confirmation, 
CoT incorrectly conflates her confirmation date (October 26) with her assumption of office (October 27), yielding an incorrect ``Conflicting Evidence/Cherry-picking'' verdict.
HerO correctly identifies the confirmation date but offers only a superficial justification, failing to address the potential ambiguity between confirmation and assumption dates. 
In contrast, DebateCV's debate structure enables nuanced interpretation: 
the Affirmative Debater supports the claim using confirmation evidence, 
while the Negative Debater challenges it by interpreting ``confirmed'' as referring to the assumption of duties. 
This exchange allows the Moderator to disambiguate the claim's true intent regarding the confirmation act itself and to issue a correct \textit{Supported} verdict.

\textbf{(2) Overlooking evidence}.
Regarding Claim 99 on New Zealand's Abortion Act, both CoT and HerO erroneously support the claim by relying solely on the rejection of proposed amendments while ignoring the Act's explicit requirement to provide medical care to a live-born infant.
In contrast, DebateCV's adversarial debate surfaces this omission: the Negative Debater identifies the overlooked statutory provision, and the Moderator correctly prioritises this direct legal evidence, yielding the accurate \textit{Refuted} verdict.

\textbf{(3) Overreliance on speculation over concrete evidence}.
For Claim 282 regarding Premier Andrews and water rights, both CoT and HerO erroneously refute the claim—CoT through subjective inference from indirect sources, and HerO by asserting ``the water was not sold to China'' despite the absence of concrete evidence about the transaction. Neither recognizes that the available evidence (e.g., trade agreements and Chinese water ownership) is insufficient to support a definitive conclusion. 
DebateCV reveals this limitation: although both Debaters resort to speculation based on indirect data, the Moderator correctly identifies the lack of direct evidence and issues the appropriate \textit{Not Enough Evidence} verdict.

\begin{figure}[t]
\centering
\begin{tikzpicture}
\begin{axis}[
    hide axis,
    xmin=0, xmax=1, ymin=0, ymax=1,
    legend style={
        at={(0.5,1)},
        anchor=north,
        legend columns=2,
        font=\small,
        draw=none,
        fill=none,
        column sep=0.5em
    },
    legend image code/.code={
        \draw[#1, draw=none, rounded corners=1pt] (0cm,-0.12cm) rectangle (0.4cm,0.12cm);
    }
]
\addlegendimage{fill=green!70!black, draw=green!50!black}
\addlegendentry{Correct}
\addlegendimage{fill=red!70!black, draw=red!50!black}
\addlegendentry{Wrong}
\end{axis}
\end{tikzpicture}

\begin{subfigure}[h]{0.48\columnwidth}
    \centering
    \begin{tikzpicture}
    \begin{axis}[debate round style]
    
    \addplot[fill=green!70!black, draw=green!50!black, bar shift=-5pt] coordinates {
        (1, 191)
        (2, 48)
        (3, 30)
    };
    
    \addplot[fill=red!70!black, draw=red!50!black, bar shift=5pt] coordinates {
        (1, 86)
        (2, 77)
        (3, 68)
    };
    
    \end{axis}
    \end{tikzpicture}
    \caption{DebateCV w/o Debate-SFT.}
    \label{fig:zero_shot}
\end{subfigure}
\hfill
\begin{subfigure}[h]{0.48\columnwidth}
    \centering
    \begin{tikzpicture}
    \begin{axis}[debate round style]
    
    \addplot[fill=green!70!black, draw=green!50!black, bar shift=-5pt] coordinates {
        (1, 261)
        (2, 94)
        (3, 11)
    };
    
    \addplot[fill=red!70!black, draw=red!50!black, bar shift=5pt] coordinates {
        (1, 69)
        (2, 57)
        (3, 8)
    };
    
    \end{axis}
    \end{tikzpicture}
    \caption{DebateCV.}
    \label{fig:full}
\end{subfigure}
\caption{Claim verification performance across rounds.}
\label{fig:combined}
\end{figure}

\begin{table}[h]
\centering
\caption{False positive rates for neutral verdicts.}
\begin{tabular}{lcccc}
\toprule
\multirow{2}{*}{\textbf{Methods}} & \multirow{2}{*}{\textbf{Acc.}} &  \multirow{2}{*}{\textbf{AVer.}} & \multicolumn{2}{c}{\textbf{False Positive Rate $\downarrow$}} \\
\cline{4-5}
& & & NEE & CEC \\
\midrule
DebateCV & \textbf{72.8} &  \textbf{54.4} & \textbf{0.4\%} & \textbf{1.1\%} \\
\quad w/o Debate-SFT & 53.8 & 41.2 & 25.0\% & 6.3\% \\
\bottomrule
\end{tabular}
\label{tab:conform}
\end{table}

\subsection{Ablation Study}\label{sec:ablation}

In this part, we evaluate ablation variants of DebateCV. 

\paragraph{\textbf{Impact of Training Data Composition.}}
We analyse two ablation variants to understand the contribution of different training data components: training without correct samples (w/o $ \mathcal{C}_{\text{correct}} $) and training without the corrected error samples (w/o $ \mathcal{C}_{\text{error}} $).

Table~\ref{tab:main} reveals complementary roles for each component. 
Removing $\mathcal{C}_{\text{error}}$ causes substantial performance drops across all conditions.
For Llama-3.1-8B, 
accuracy decreases from 83.4\% to 76.4\% with Golden evidence,
from 72.8\% to 63.0\% under Retrieved (H),
and from 73.6\% to 68.2\% with Retrieved (I).
Qwen-2.5-7B exhibits similar degradation patterns.
Removing $\mathcal{C}_{\text{correct}}$, however, shows divergent effects depending on the evidence condition: 
it preserves competitive performance when using retrieved evidence but incurs more substantial performance degradation under golden evidence. For instance, Llama-3.1-8B attains 70.8\% accuracy under Retrieved (H), which is only 2.4\% below that of the full model. 
Similarly, its AVeriTeC score also remains almost unaffected (dropping only 0.4-1.6\% in all setups). 
In contrast, its accuracy drops more markedly under golden evidence, 
falling to 76.8\% from the full model's 83.4\%.
The results show that \textbf{the correct samples $\mathcal{C}_{\text{correct}}$
are only important for high-quality evidence scenarios,
while corrected samples $\mathcal{C}_{\text{error}}$ are crucial for robust performance under noisy evidence conditions, highlighting the importance of our error correction mechanism.}
Incorporating both data components achieves optimal performance across all conditions.

\paragraph{\textbf{Impact of Debate-SFT}}  

As shown in Table~\ref{tab:main}, our Debate-SFT method yields substantial improvements in the Moderator's performance compared to its zero-shot counterpart.
For instance, under Retrieved (H), the w/o Debate-SFT variant exhibits significant performance degradation, with accuracy dropping from 72.8\% to 53.8\% when using Llama-3.1-8B and from 72.6\% to 54.2\% when using Qwen-2.5-7B.
The results highlight the risks of using zero-shot LLM agents, such as those in \cite{han2025beyond}, for this task.

To further interpret these improvements, 
Fig.~\ref{fig:combined} compares DebateCV performance with and without Debate-SFT, using Llama-3.1-8B as Moderators across debated rounds under the Retrieved (H) condition.
It is observed that the zero-shot Moderator resolves most claims in the first debate round, but its error rate rises in subsequent rounds.
This is likely due to the inherent complexity of claims requiring extended debate, compounded by known LLM performance degradation in multi-turn dialogues in various tasks~\cite{llm_lost,tcss_stance,icdm_stance}. 
In contrast, the Moderator post-trained by Debate-SFT sustains higher accuracy across all rounds, 
which shows \textbf{enhanced decision-making for both simple claims resolvable in a single round and more challenging ones}.

Further, we observe that the zero-shot Moderator frequently defaults to neutral verdicts, i.e., \textit{Not Enough Evidence} (NEE) and \textit{Conflicting Evidence/Cherry-picking} (CEC), 
even when sufficient evidence exists to definitively support or refute a claim.
This may stem from conformity bias in LLMs~\cite{conform1, conform2}, where the Moderator avoids contradicting either Debater. 
For example, in Table~\ref{tab:conform}, the zero-shot Moderator yields high false positive rates of 25.0\% for NEE and 6.3\% for CEC. 
Nonetheless, Debate-SFT dramatically reduces the false positive rates to 0.4\% for NEE and 1.1\% for CEC, respectively, 
highlighting its benefits in \textbf{mitigating unwarranted neutrality and generating decisive verdicts}.

\subsection{Human Evaluation on Synthetic Data}\label{sec:validation}

To validate the synthetic data in the SynDeC dataset, i.e., the debate recordings and the corrected justifications for \(\mathcal{C}_{\text{error}}\),
we recruited the same participants in the human evaluation in Sec.~\ref{sec:main} to evaluate the quality of the synthetic debate recordings and justifications for 50 randomly selected samples. 
Following the evaluation protocol of \citet{nan2024let}, 
we asked the participants to rate each sample on a three-point Likert scale (``\textit{not likely},'' ``\textit{likely},'' or ``\textit{very likely}'') in response to the following questions:
\begin{enumerate}
\item Question 1 (for Debates):
``\textit{To what extent do the Debaters consistently defend their assigned stance with relevant evidence and generate a well-reasoned argument?}''
\item Question 2 (for Justifications):
``\textit{To what extent does this justification accurately summarize and reflect the debate?}''
\end{enumerate}
Each sample was assessed by at least three participants.

\begin{figure}[h]
    \centering
    \begin{tikzpicture}
    \definecolor{oiOrange}{HTML}{E69F00}
    \definecolor{oiSky}{HTML}{56B4E9}
    \definecolor{oiGreen}{HTML}{009E73}
    \begin{axis}[
        ybar stacked,
        bar width=24pt,
        width=0.8\columnwidth,
        height=4.3cm,
        symbolic x coords={Debate, Justification},
        xtick=data,
        ymin=0,
        ymax=100,
        ytick={0,20,40,60,80,100},
        yticklabel={\pgfmathprintnumber{\tick}\,\%},
        axis lines=left,
        enlarge x limits=0.5,
        ymajorgrids,
        grid style={gray!20},
        tick style={black},
        tick label style={font=\small, color=black},
        label style={font=\small},
        legend style={font=\small, at={(0.5,1.02)}, anchor=south, legend columns=3, /tikz/every even column/.append style={column sep=6pt}, text=black, draw=none},
        legend image code/.code={\draw[#1, draw=none] (0cm,-0.09cm) rectangle (0.32cm,0.09cm);},
        nodes near coords,
        nodes near coords style={font=\footnotesize, text=black},
        nodes near coords align={vertical},
        nodes near coords={\pgfmathprintnumber{\pgfplotspointmeta}\%}
    ]
    \addplot+[fill=oiOrange, draw=none] coordinates {(Debate,3.8) (Justification, 5.2)};
    \addplot+[fill=oiSky, draw=none] coordinates {(Debate, 52.2) (Justification, 51.1)};
    \addplot+[fill=oiGreen, draw=none] coordinates {(Debate, 44.0) (Justification, 43.7)};
    \legend{Not Likely, Likely, Very Likely}
    \end{axis}
    \end{tikzpicture}
    \caption{Human evaluation results on the quality of the generated debates and justifications.}
    \label{fig:generated_ratings}
\end{figure}

As shown in Fig.~\ref{fig:generated_ratings}, both the generated debates and justifications received high ratings: 96.2\% of debates were rated as ``\textit{likely}'' or ``\textit{very likely}'' for Question 1, and over 94.8\% of justifications received the same ratings for Question 2. 
Inter-annotator agreement was high, with Krippendorff's $\alpha = 0.83$ for debate quality (excellent) and $\alpha = 0.71$ for justification quality (substantial), respectively.
These results indicate that \textbf{our Debaters consistently generate substantive discussions to support their reasoning, and that the resulting justifications effectively distil the arguments presented in the debates, all without hallucinations}.

\subsection{Impact of LLMs for Debaters}\label{sec:llm_impact}
DebateCV addresses claim verification through debates among Debaters and a Moderator.
In Sec.~\ref{sec:main},
we have validated that the Moderator in our DebateCV framework is robust to different LLM backbones.
Here,
we further analyse the impact of using different LLM backbones for the Debaters on the overall performance.

\begin{table}[h]
\centering
\caption{Impact of different LLMs for Debaters on DebateCV performance under Retrieved (H) condition.}
\label{tab:debater}
\begin{tabular}{lcc}
\toprule
\textbf{Debater LLM} & \textbf{Acc.} & \textbf{AVer.} \\
\midrule
GPT-4o-Mini   & 72.8 & 54.4 \\
Llama-3.1-8B  & 72.6 & 54.2 \\
Qwen-2.5-7B   & 71.8 & 53.8 \\
\bottomrule
\end{tabular}
\end{table}

Table~\ref{tab:debater} shows the performance of DebateCV under the Retrieved (H) condition, where the Moderator is fixed as post-trained Llama-3.1-8B and the Debater is set to either GPT-4o-Mini, Llama-3.1-8B, or Qwen-2.5-7B.
The results show that the choice of Debater model has only a minor effect on accuracy and AVeriTeC score, 
with variations remaining within a narrow range (from 72.8 to 71.8 for accuracy and from 54.4 to 53.8 for AVeriTeC score). Thus, \textbf{DebateCV is robust to different LLM backbones for the Debaters.}

\section{Conclusion}
In this work, we introduce DebateCV, the first debate-driven claim verification framework. DebateCV orchestrates multiple LLM agents in structured adversarial debates, where two Debaters take opposing stances and present evidence-grounded arguments, while a Moderator synthesises these debates into final verdicts.
To improve the Moderator's ability to adjudicate complex, multi-round debates, we propose Debate-SFT, a novel post-training framework that leverages synthetic debate data generated through a three-stage pipeline: debate data synthesis, error correction, and post-training.
Extensive experiments on the state-of-the-art AVeriTeC benchmark show that DebateCV significantly outperforms both single-agent and multi-agent baselines across a range of evidence quality conditions. 
Our analysis reveals that these performance gains stem from DebateCV's adversarial cross-validation mechanism to mitigate three fundamental error patterns prevalent in existing methods: claim/evidence misinterpretation, overlooking relevant evidence, and overreliance on speculation.
Beyond quantitative improvements, human expert evaluations show that DebateCV produces substantially higher-quality justifications, enhancing the transparency and trustworthiness essential for practical fact-checking applications. 
These technical advancements will enhance digital literacy and strengthen misinformation countermeasures with high-quality justifications, thereby fostering a more trustworthy information ecosystem.
Future research could explore methods to enhance the inference cost efficiency of DebateCV,\footnote{Appendix~\ref{sec:cost} analysed the inference cost of DebateCV.} as well as strategies to improve claim verification performance under extremely noisy or even heavily polluted evidence conditions.

\section*{Acknowledgments}
This work was done while Haorui He was under the supervision of Yupeng Li. This work was supported by National Natural Science Foundation of China (No. 62202402),
Guangdong and Hong Kong Universities “1+1+1” Joint Research Collaboration Scheme, Project No. 2025A0505000001,
the Early Career Scheme (ECS) from the Research Grants Council of HKSAR (HKBU 22202423),
the General Research Fund (GRF) from the Research Grants Council of HKSAR (HKBU 12203425),
a grant from the Germany/Hong Kong Joint Research Scheme sponsored by the Research Grants Council of HKSAR and the German Academic Exchange Service of Germany (No. G-HKBU208/25),
the Initiation Grant for Faculty Niche Research Areas 2023/24 (No. RC-FNRA-IG/23-24/COMM/01),
Research Cluster Matching Scheme (No. RCMS/24-25/01) of Hong Kong Baptist University,
National Natural Science Foundation of China (No. 62072115),
the Shanghai Science and Technology Innovation Action Plan Project (No. 22510713600),
the grants from the Research Grants Council of HKSAR (HKU 17202325),
the University of Hong Kong (Project 2409100399),
the HKU Faculty Exchange Award 2024 (Faculty of Engineering), 
Scientific Research Innovation Capability Support Project for Young Faculty (No. SRICSPYF-BS2025137), and Startup Grant (Tier 1) for New Academics AY2020/21 of Hong Kong Baptist University.
\bibliography{main}
\bibliographystyle{ACM-Reference-Format}
\newpage
\input{appendix}

\end{document}

%% file: appendix.tex
\appendix
\section{Definitions of the Verdict Categories}\label{sec:definition}
Similar to \cite{averitec, infact, hero}, we consider four categories of verdicts for the claims.
Below is a summary of their definitions.
(i) \textbf{Supported}: The claim is supported by the arguments and evidence presented.  
(ii) \textbf{Refuted}: The claim is contradicted by the arguments and evidence presented.
(iii) \textbf{Not Enough Evidence}: 
    The presented evidence is not enough to support or refute the claim.
    This category applies when the evidence either explicitly indicates that relevant evidence cannot be found or leaves certain aspects of the claim neither supported nor refuted. 
(iv) \textbf{Conflicting Evidence/Cherry-picking}: The claim is misleading due to conflicting evidence or cherry-picking, but is not explicitly refuted. 
    This category includes cases such as cherry-picking (selectively presenting evidence to misrepresent truth), true-but-misleading (e.g., ``\texttt{Alice has never lost an election}'' when Alice has only ever run unopposed), and instances where contradictory evidence is found.
Please see \citep{averitec} for a detailed description of these categories and examples.

\section{Prompts}\label{sec:prompt}
This section presents the prompts utilized in our work to direct the behavior of the LLM agents: the Debaters, Moderator, and Corrector.

\subsection{Pre-debate Configuration Prompts}
In pre-debate configuration, meta-prompts are provided to both Debaters and the Moderator to instruct their actions throughout the multi-round debate.
The meta-prompt for \textit{Debaters} is as follows:
\begin{calloutblock}
You are a Debater in a fact-checking scenario. Your task is to verify the accuracy of a specific claim by either affirming or refuting it based on the provided credible evidence.

\textbf{Your Role}:\\
- Defend your assigned position using credible evidence.\\
- Critically analyze and challenge the opposing argument respectfully and factually.

\textbf{Objective}:\\
- Determine the truthfulness of the claim.\\
- Provide compelling evidence to support your stance.\\
- Critique the opposing evidence to strengthen your position, while being concise.

\textbf{Claim}: [CLAIM]

\textbf{Evidence Set}: [EVIDENCE\_SET]
\end{calloutblock}

The meta-prompt for \textit{Moderator} is as follows:
\begin{calloutblock}
You are the Moderator in a fact-checking debate where two Debaters examine the truthfulness of a given claim by presenting credible supporting or opposing evidence. Your goal is to facilitate a fact-based evaluation of the claim, ensuring each side maintains a strong, evidence-backed stance.

\textbf{Responsibilities}:\\
- Guide each debate round, ensuring arguments remain evidence-based.\\
- Assess the relevance and strength of the credible evidence presented by both sides.\\
- Determine if further rounds are essential based on the new insights provided. End the debate if both sides are just repeating their previous arguments without bringing new insights.

\textbf{Claim}: [CLAIM]

\textbf{Evidence Set}: [EVIDENCE\_SET]

\textbf{Verdict Criteria}:\\
- Supported: The claim is fully supported by the credible evidence provided.\\
- Refuted: The claim is directly contradicted by the credible evidence.\\
- Not Enough Evidence: Insufficient credible evidence to confirm or refute the claim.\\
- Conflicting Evidence/Cherry-picking: The claim is misleading due to conflicting evidence or selective information.

\end{calloutblock}

\subsection{In-debate Process Prompts}
After the pre-debate configuration, we use the following prompt to instruct the \textit{Affirmative Debater} to initialize the debate.
\begin{calloutblock}
\textbf{Claim}: [CLAIM]

You are affirming the veracity of the claim. Use the credible evidence provided to construct a compelling argument supporting the claim.

\textbf{Approach}:\\
- Decompose the claim into its core components and clearly articulate its meaning.\\
- Present the most relevant evidence to substantiate your position, citing each piece as (Content of evidence, Source URL).\\
- Critically analyze and refute the opposition's points using credible evidence (this can be omitted in the first round).
\end{calloutblock}

The \textit{Negative Debater} is guided to provide a rebuttal:
\begin{calloutblock}
\textbf{Previous Argument (Affirmative)}: [AFFIRMATIVE\_ARGUMENT]

You are refuting the claim, arguing that it is false. Use the credible evidence provided to construct a compelling argument against the claim.

\textbf{Approach}:\\
- Present the most relevant evidence to support your position, citing each piece as (Content of evidence, Source URL).\\
- Critically analyze and counter the opposition's points using credible evidence.
\end{calloutblock}

\begin{table*}[h]
\centering
\caption{Examples of DebateCV outputs for claim verification. Accurate judgments are in \good{green}.}
\small
\resizebox{\textwidth}{!}{%
\begin{tabular}
{p{0.31\textwidth}p{0.33\textwidth}p{0.33\textwidth}p{0.3\textwidth}}
\midrule
\multirow{2}{*}{\centering \textbf{Claim (ID; Label)}} &
\multicolumn{3}{c}{\textbf{DebateCV}} \\
\cline{2-4}
 & \textbf{Affirmative Debater} & \textbf{Negative Debater} &  \textbf{Moderator}  \\
\midrule
UNESCO declared Nadar community as the most ancient race in the world.
\textbf{(ID: 3; Refuted)}
&
The Nadar community has a rich historical presence and cultural significance. Some scholars recognize their deep roots. Supported.
&
No direct recognition is from UNESCO. Many communities have ancient lineages, and cultural prominence does not equate to ``most ancient.'' Refuted.
&
\good{The Negative effectively refutes the Affirmative by highlighting its lack of official UNESCO endorsement.} Refuted. 
\\
\midrule
The USA and its Western allies have been using their media outlets to publish articles based on fabricated information under allegations of non-compliance with the Chemical Weapons Convention.
\textbf{(ID: 5; Refuted)}
&
There are accusations from multiple sources—including pro-Kremlin and Western outlets—that narratives around Syria's compliance were manipulated. OPCW reports have also been contested. Supported.
&
OPCW investigations consistently raised concerns about Syria's compliance, supported by verified findings. Reports of chemical use are corroborated by independent missions. Allegations are based on real evidence. Refuted.
&
\good{The Negative side provided stronger and more concrete evidence, including verified OPCW findings.} Affirmative claims lacked proof of systemic fabrication. Refuted. 
\\
\midrule
``Why should you pay more taxes than Donald Trump pays?'' Biden cited \$750 paid by Trump in 2016 and 2017, and quoted Trump saying, ``Because I'm smart. I know how to game the system.''
\textbf{(ID: 7; Supported)}
&
Multiple reputable sources (NYT, NPR, Forbes) confirm Trump paid only \$750 in federal income tax during those years. His own quote suggests he exploited legal loopholes. Supported.
&
Breitbart disputes the report, suggesting it misrepresents Trump's filings. Not all wealthy individuals avoid taxes, but many legally reduce liabilities through established strategies. The claim oversimplifies tax complexities. Refuted.
&
Despite counterarguments, the Affirmative side presented credible and consistent evidence from multiple reputable sources. \good{The Negative side failed to refute the core facts.} Supported. 
\\
\midrule
Joe Biden wants to ban fracking.
\textbf{(ID: 23; Refuted)}
&
Biden supports banning new oil and gas permits on public lands, which limits fracking. While he hasn't explicitly called for a ban, his campaign indicated restrictions that align with a de facto ban. Supported.
&
Biden explicitly stated he does not support a complete ban, instead he aims to restrict new permits on public lands. Campaign statements clarify this nuance. Refuted.
&
\good{The Negative correctly identifies Biden's nuanced stance.} His policies target new permits, not an outright ban. Refuted. \\
\midrule
The New York Post endorsed Donald Trump's bid for re-election.
\textbf{(ID: 32; Supported)}
&
The NYP editorial board explicitly endorsed Trump in the 2020 election, despite some policy criticisms, which reflects strategic alignment rather than full agreement. Supported.
&
While the Post has conservative leanings, its endorsement included reservations and critiques, suggesting a complex rather than clear-cut stance. Refuted.
&
\good{The Affirmative presents direct evidence of the endorsement}, while the Negative raises contextual concerns without disproving it. Supported. 
\\
\midrule
The COVID-19 pandemic was pre-planned with the help of Bill Gates.
\textbf{(ID: 35; Refuted)}
&
Gates has long warned of pandemics, which suggests foreknowledge and planning, fueling public perception of involvement. Supported.
&
The advocacy was for preparedness, not prediction. Conspiracy theories lack factual basis. Refuted.
&
\good{Affirmative relies on speculation}, while the Negative clearly distinguishes between warning and orchestration. Refuted. 
\\
\midrule

Breitbart News reports that the daughter of Delaware Democratic Senator Chris and seven other underage girls were featured on Hunter Biden's laptop.
\textbf{(ID: 37; Refuted)}
&
Allegations from political figures like Lauren Witzke suggest there may be merit to the claim. Lack of denial implies some level of credibility. Supported.
&
No credible source confirms Breitbart reported this. Fact-checks disprove child pornography claims. Unsupported allegations should not be treated as fact. Refuted.
&
\good{Affirmative relies on unverified allegations}, while the Negative shows absence of actual reporting or evidence. Refuted.
\\
\midrule
The umbilical cord should not be cut until 1 hour after birth or the baby will not have enough blood right \textbf{(ID: 46; Refuted)}
&
ACOG recommends delayed clamping (30–60 sec) to improve hemoglobin and iron levels. A 1-hour delay may offer benefits. Supported.
&
ACOG supports only 30–60 seconds of delay. Longer delay lacks evidence and may not be practical. Refuted.
&
\good{Delayed clamping is beneficial, but notes no support for 1-hour delay.} Medical guidelines back shorter window. Refuted. 
\\
\midrule
People who do not vote for the BJP in the 2020 elections will not get the COVID vaccine free of charge.
\textbf{(ID: 51; Refuted)}
&
BJP manifesto promised free vaccines, but critics feared this might be tied to voter loyalty. Political context implied exclusion. Supported.
&
Official statements clarified vaccines would be available to all priority groups regardless of political affiliation. Refuted.
&
No evidence \good{linking eligibility to voting behavior.} Government immunization programs apply universally. Refuted. 
\\
\midrule
Officer who wore Trump 2020 mask to polls to face disciplinary action.
\textbf{(ID: 63; Supported)}
&
A police officer was photographed wearing a 'Trump 2020' mask at a polling location. Mayor Suarez affirmed consequences. Supported.
&
While the officer likely wore the mask and may face discipline, the term 'disciplinary action' is vague. Free speech rights and political expression complicate the issue. Refuted.
&
Multiple evidences confirm both the mask-wearing and disciplinary action. \good{The Negative raises valid questions but doesn't contradict the central claim.} Supported.  
\\
\bottomrule
\end{tabular}%
}
\label{tab:examples}
\end{table*}

At the end of each round, the \textit{Moderator} oversees the debate:

\begin{calloutblock}
Round [ROUND\_NUMBER] of the fact-checking debate has concluded.

\textbf{Affirmative}: [AFFIRMATIVE ARGUMENT]

\textbf{Negative}: [NEGATIVE ARGUMENT]

As the Moderator, evaluate each side's arguments by examining the relevance and sufficiency of the evidence presented, while being concise.

\textbf{Steps to Follow}:\\
1. Summarize the new insights from this round compared to previous rounds.\\
2. Note any missing evidence or arguments in either side's case.\\
3. Assess if further debate is necessary or if the arguments are repeating previous points without adding substantial new information.\\
4. Conclusion:\\
- If a clear verdict is supported or there is no need for further debate: Provide justification for this outcome; Select one of the following verdict labels: "Supported", "Refuted", "Not Enough Evidence", or "Conflicting Evidence/Cherry-picking"; Set "Proceeding Necessity" to "No".\\
- If further debate is essential: Indicate why additional rounds are necessary; Set "Proceeding Necessity" to "Yes"; Provide a "Justification for Proceeding" outlining the rationale for needing further evidence; Leave "Justification for Verdict" and "Verdict" blank.
Output your findings in JSON format:
\{"Primary Insight": "...",
  "Evidence Gaps": "...",
  "Justification for Proceeding": "...",
  "Proceeding Necessity": "Yes/No",
  "Justification for Verdict": "...",
  "Verdict": "Supported", "Refuted", "Not Enough Evidence", or "Conflicting Evidence/Cherry-picking"\}

\end{calloutblock}

If the Moderator opts to continue the debate, the following interaction prompt will be provided to the \textit{Affirmative Debater}.

\begin{calloutblock}
\textbf{Opposition Argument}: [OPPOSITION ARGUMENT]

Do you agree with this perspective? Provide your response explaining your reasoning, supporting evidence, and highlighting any weaknesses in the opposition.
\end{calloutblock}

If the Moderator determines that the debate has converged, the debate will be terminated.
In cases where the debate lasts for more than three rounds, the \emph{Moderator} receives the final prompt to output a verdict and generate a corresponding justification.

\begin{calloutblock}

Affirmative: [AFFIRMATIVE\_ARGUMENT]
Negative: [NEGATIVE\_ARGUMENT]

Summarize the primary insights gathered throughout the entire debate concisely.

After reviewing both sides' arguments on the claim:
[CLAIM]

Select a verdict from the following labels based on the credible evidence: "Supported", "Refuted", "Not Enough Evidence", or "Conflicting Evidence/Cherry-picking".

Provide a step-by-step justification for your choice, then present your conclusion in JSON format:
\{
  "Justification for Verdict": "...",
  "Verdict": "Supported", "Refuted", "Not Enough Evidence", or "Conflicting Evidence/Cherry-picking"
\}
\end{calloutblock}

\subsection{Corrector's Prompt}
\emph{Corrector} generates a justification explaining how the ground-truth verdict can be derived from the debate.
\begin{calloutblock}
Debate Recording: [DEBATE\_RECORDING]

Primary Insights: [Primary\_Insights]

Task: Please provide the justification for the verdict that this claim is [GT\_VERDICT] based on the debate context. 

The output should be in JSON format. 
\{
"Justification for Verdict": "..."
\}
\end{calloutblock}

\section{Examples of DebateCV Outputs}\label{sec:examples}
To demonstrate how DebateCV performs claim verification, we provide ten examples in Table~\ref{tab:examples}, in addition to the case study detailed in Section~\ref{sec:experiment}. Each example consists of an input claim, followed by arguments presented by the Affirmative and Negative Debaters, and concludes with the Moderator's final decision. Agent responses are condensed across all debate rounds for brevity.

\section{Computational Cost Analysis}\label{sec:cost}
\begin{table}[htbp]
\centering
\caption{Computational cost per claim verification.}
\resizebox{\linewidth}{!}{%
\begin{tabular}{cccccc}
\toprule
\multirow{2}{*}{\textbf{Role}} & 
\multicolumn{2}{c}{\textbf{Input}} & 
\multicolumn{2}{c}{\textbf{Output}} & 
\multirow{2}{*}{\makecell{\textbf{Total} \\ \textbf{Cost (\$)}}} \\
\cmidrule(lr){2-3} \cmidrule(lr){4-5}
& Tokens & Cost (\$) & Tokens & Cost (\$) & \\
\midrule
Debaters (GPT-4o-Mini) & 3685.13 & 0.0005 & 1526.01 & 0.0009 & 0.0014 \\
Moderator (Qwen-2.5-7B) & 3355.58 & 0.0006 & 290.78 & 0.0002 & 0.0008 \\
\cmidrule(lr){1-6}
\quad \textbf{Total} & 7040.71 & 0.0011 & 1816.79 & 0.0011 & 0.0022 \\
\bottomrule
\end{tabular}%
}
\label{tab:cost_analysis}
\end{table}

DebateCV may incur higher computational costs than single-agent baselines due to multi-round interactions among agents. 
Here, we analyse its inference cost under the Retrieved (H) condition.
In DebateCV, Debaters are powered by GPT-4o-Mini (OpenAI API: \$0.15/1M input tokens, \$0.60/1M output tokens). 
The Moderator uses a post-trained Llama-3.1-8B or Qwen-2.5-7B model; we estimate its cost with the official Qwen-2.5-7B API (Alibaba Cloud: \$0.175/1M input tokens, \$0.70/1M output tokens).

As shown in Table~\ref{tab:cost_analysis}, verifying one claim costs \$0.0014 for Debaters and \$0.0008 for the Moderator, totaling \$0.0022.
For context, PolitiFact published 74 fact-check reports from July 1 to September 30, 2024, as counted on their official website. 
Using DebateCV would cost $74 \times 0.0022 = \$0.1628$, which is obviously more cost-efficient than human fact-checkers.
While we acknowledge that DebateCV can be relatively more expensive than single-agent methods like HerO, 
this additional cost is common in multi-agent frameworks \cite{debate_improve_factuality,mad} 
and is justified by the significant accuracy gains, especially in high-stakes applications like fact-checking, 
where even marginal improvements can yield substantial real-world impact.